\documentclass[runningheads]{llncs}
\usepackage{graphicx}
\usepackage{amsmath,amssymb} 
\usepackage{color}

\usepackage{wasysym}
\newcommand*{\vimage}[1]{\vcenter{\hbox{\includegraphics[width=1.6cm]{#1}}}}
\newcommand*{\vpointer}{\vcenter{\hbox{\scalebox{1.8}{\Huge\pointer}}}}
\usepackage{multirow}
\usepackage{capt-of}

\begin{document}
\title{Reinforced Temporal Attention and Split-Rate Transfer for Depth-Based Person Re-Identification} 

\titlerunning{Reinforced Temporal Attention for Depth-Based Person Re-Identification}
%
\author{Nikolaos Karianakis\inst{1} \and
Zicheng Liu\inst{1} \and
Yinpeng Chen\inst{1} \and
Stefano Soatto\inst{2}}
%
\authorrunning{N. Karianakis, Z. Liu, Y. Chen and S. Soatto}
%
\institute{Microsoft, Redmond, USA \and
University of California, Los Angeles, USA}
\maketitle              
\begin{abstract}
We address the problem of person re-identification from commodity depth sensors. 
One challenge for depth-based recognition is data scarcity. Our \emph{first} contribution addresses this problem 
by introducing \emph{split-rate} RGB-to-Depth transfer, which leverages large RGB datasets more effectively 
than popular fine-tuning approaches. Our transfer scheme is based on the observation that the model parameters 
at the bottom layers of a deep convolutional neural network can be directly shared between RGB and depth data 
while the remaining layers need to be fine-tuned rapidly. Our \emph{second} contribution enhances re-identification for video 
by implementing temporal attention as a Bernoulli-Sigmoid unit acting upon frame-level features. Since this unit 
is stochastic, the temporal attention parameters are trained using reinforcement learning. 
Extensive experiments validate the accuracy of our method in person re-identification from depth sequences. 
Finally, in a scenario where subjects wear unseen clothes, we show large performance gains compared to 
a state-of-the-art model which relies on RGB data.
\keywords{Person Re-Identification from Depth \and Reinforced Temporal Attention \and Split-Rate Transfer.}
\end{abstract}
\section{Introduction}
\label{intro}

Person re-identification is a fundamental problem in automated video surveillance and has attracted significant attention 
in recent years~\cite{galaS14,vezzaniBC13,gongCYL14}. When a person is captured by cameras with non-overlapping views, 
or by the same camera but over many days, the objective is to recognize them across views among a large number of imposters. 
This is a difficult problem because of the visual ambiguity in a person's appearance due to large variations in illumination, human pose, 
camera settings and viewpoint. Additionally, re-identification systems have to be robust to partial occlusions and cluttered background. 
Multi-person association has wide applicability and utility in areas such as robotics, multimedia, forensics, autonomous driving and cashier-free shopping.

\begin{figure}[t]
\begin{minipage}{.5\linewidth}
 \begin{center}
  \begin{minipage}{.3\columnwidth}
    \hspace{-0.6mm}
    \includegraphics[width=1.6cm]{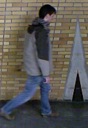}
  \end{minipage}%
  \begin{minipage}{.7\columnwidth}
    \hspace{-2.3mm}
    \includegraphics[width=4.3cm]{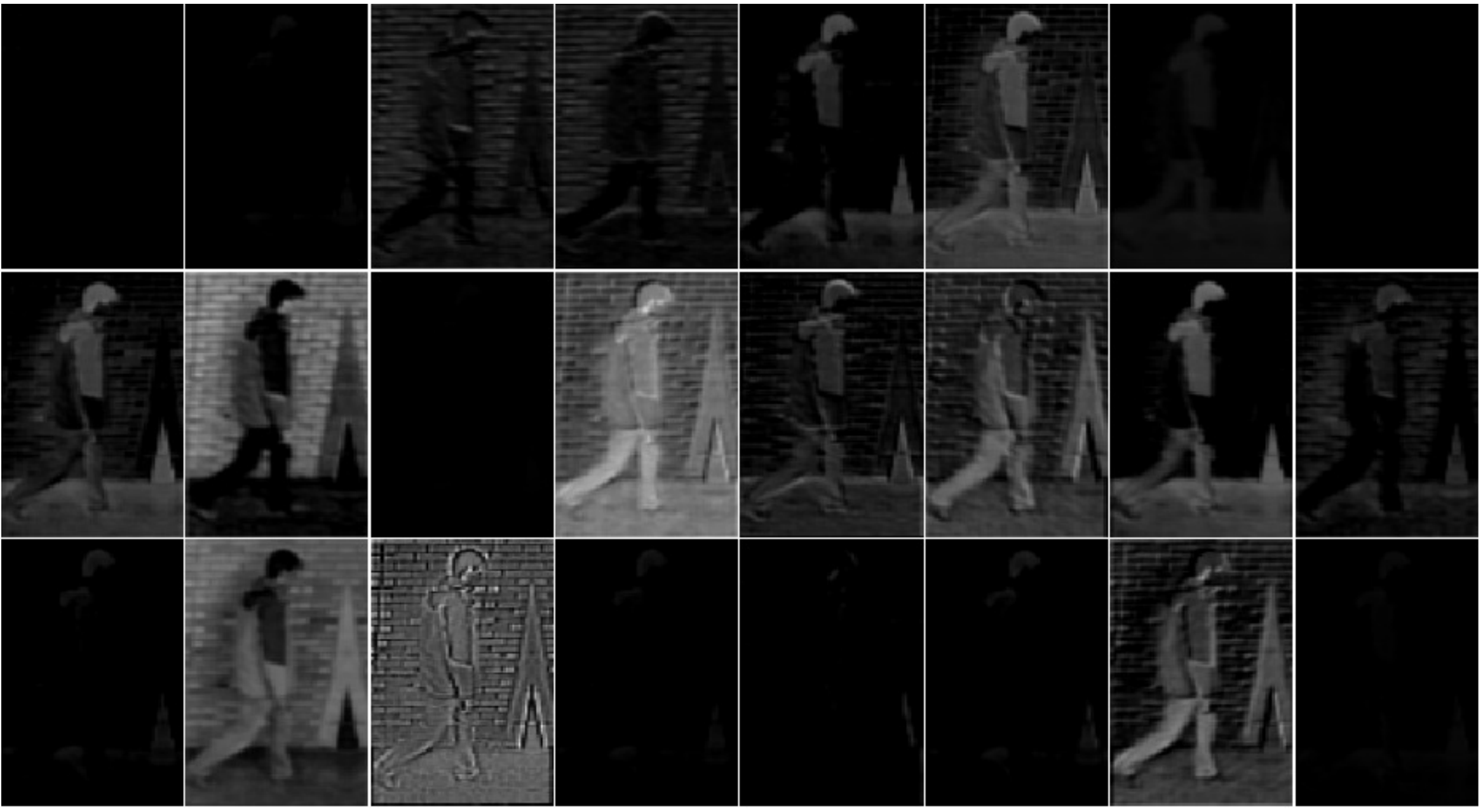}
  \end{minipage}
  \begin{minipage}{.5\columnwidth}
    \centering
    \includegraphics[width=2.95cm]{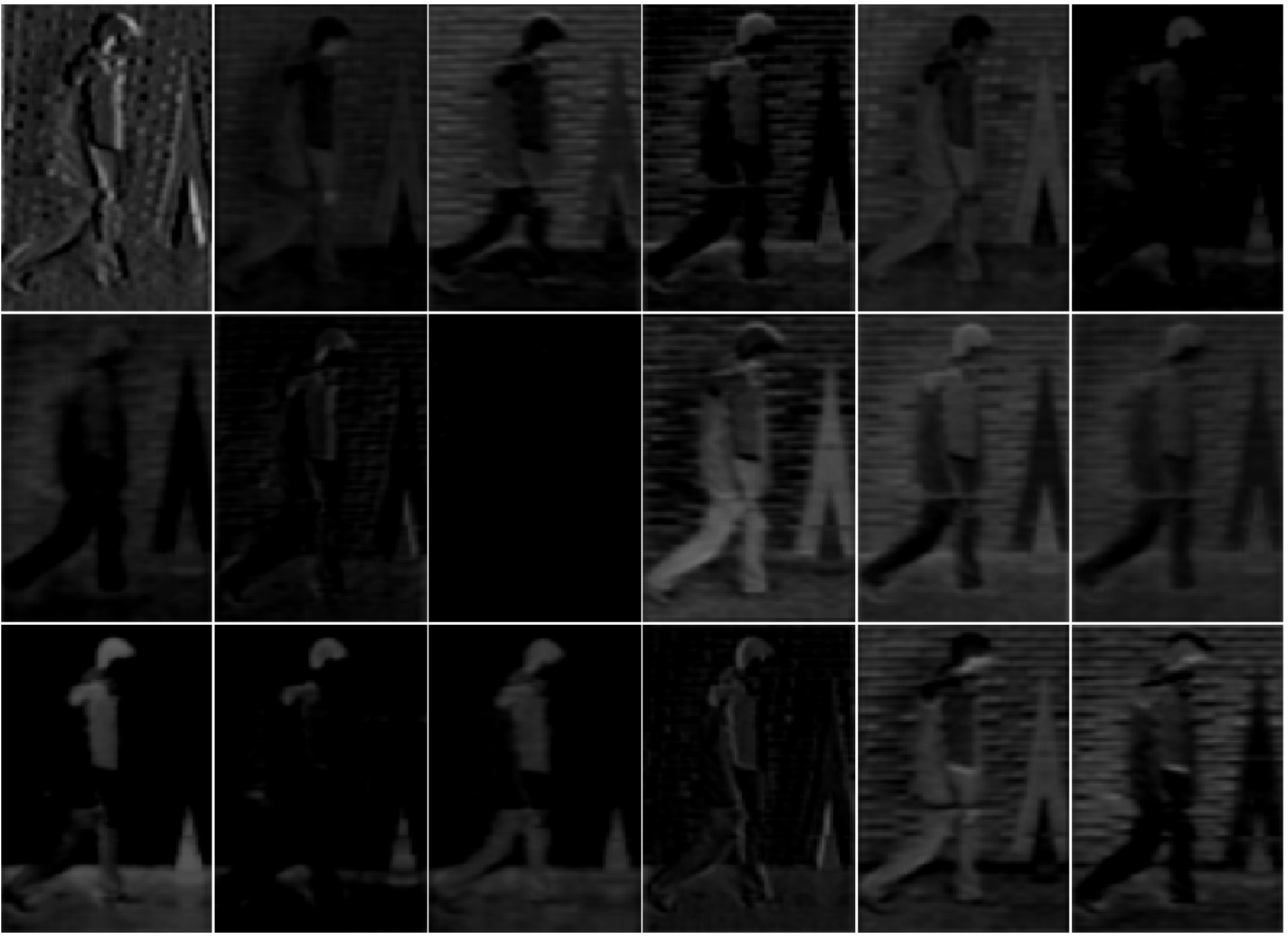}
  \end{minipage}%
  \begin{minipage}{.5\columnwidth}
    \includegraphics[width=2.95cm]{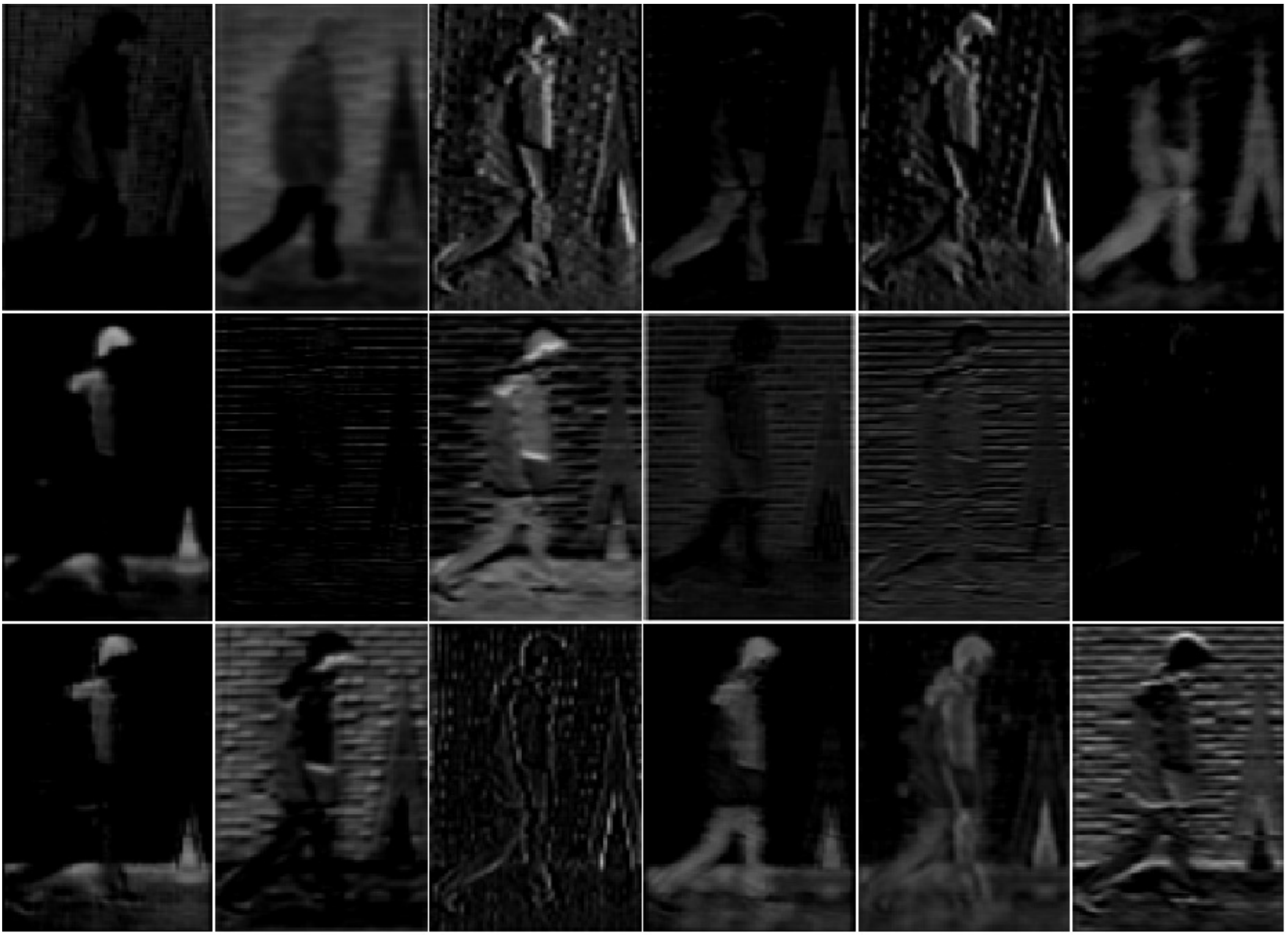}
  \end{minipage}
\vspace{1mm}
\\(a) Person ReID from RGB~\cite{xiaoLOW16}
 \end{center}
\end{minipage}%
\hfill
\begin{minipage}{.5\linewidth}
 \begin{center}
  \begin{minipage}{.3\columnwidth}
    \hspace{-0.2mm}
    \includegraphics[width=1.6cm]{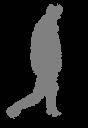}
  \end{minipage}%
  \begin{minipage}{.7\columnwidth}
     \hspace{-2mm}
    \includegraphics[width=4.3cm]{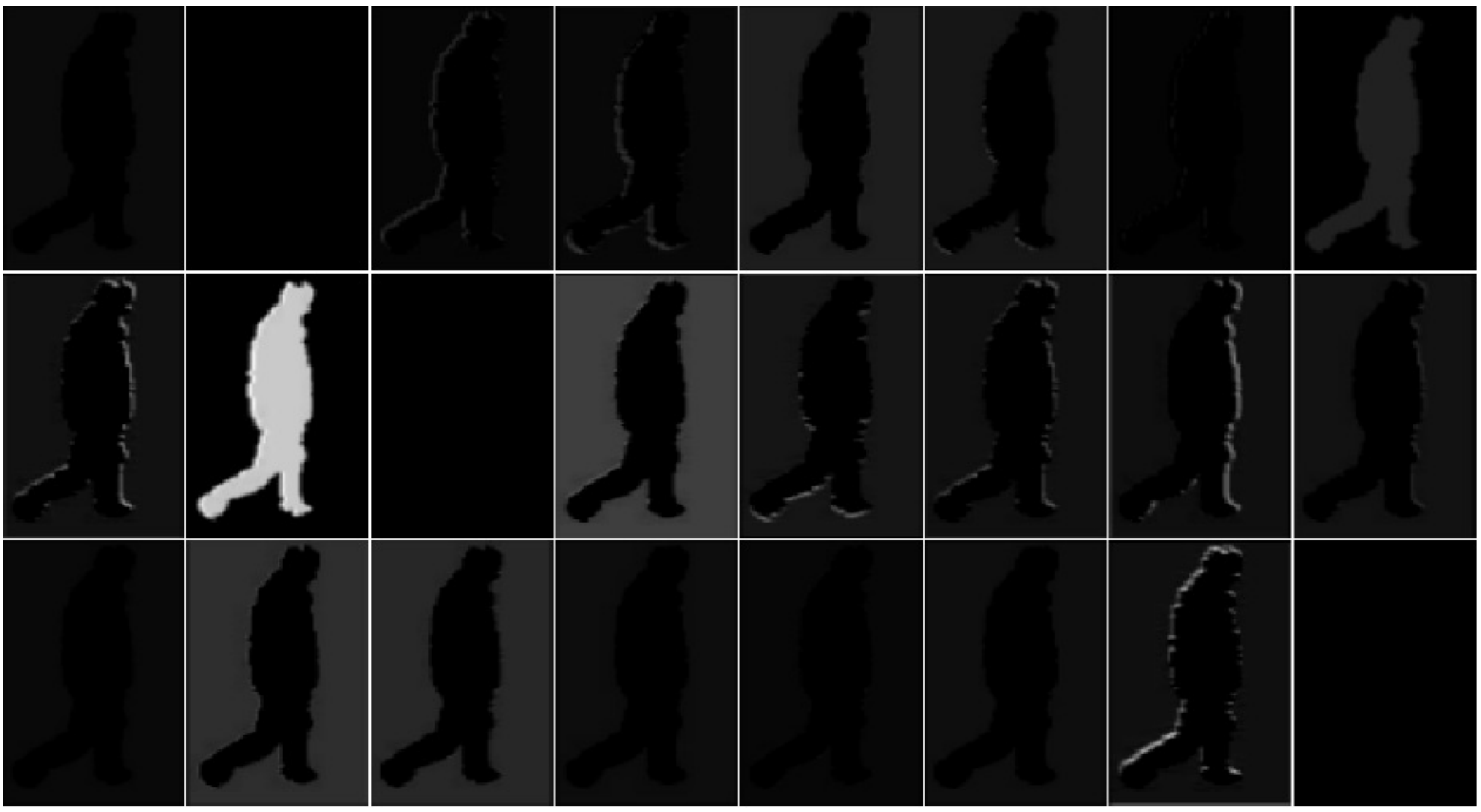}
  \end{minipage}
  \begin{minipage}{.5\columnwidth}
    \hspace{-0.2mm}
    \includegraphics[width=2.95cm]{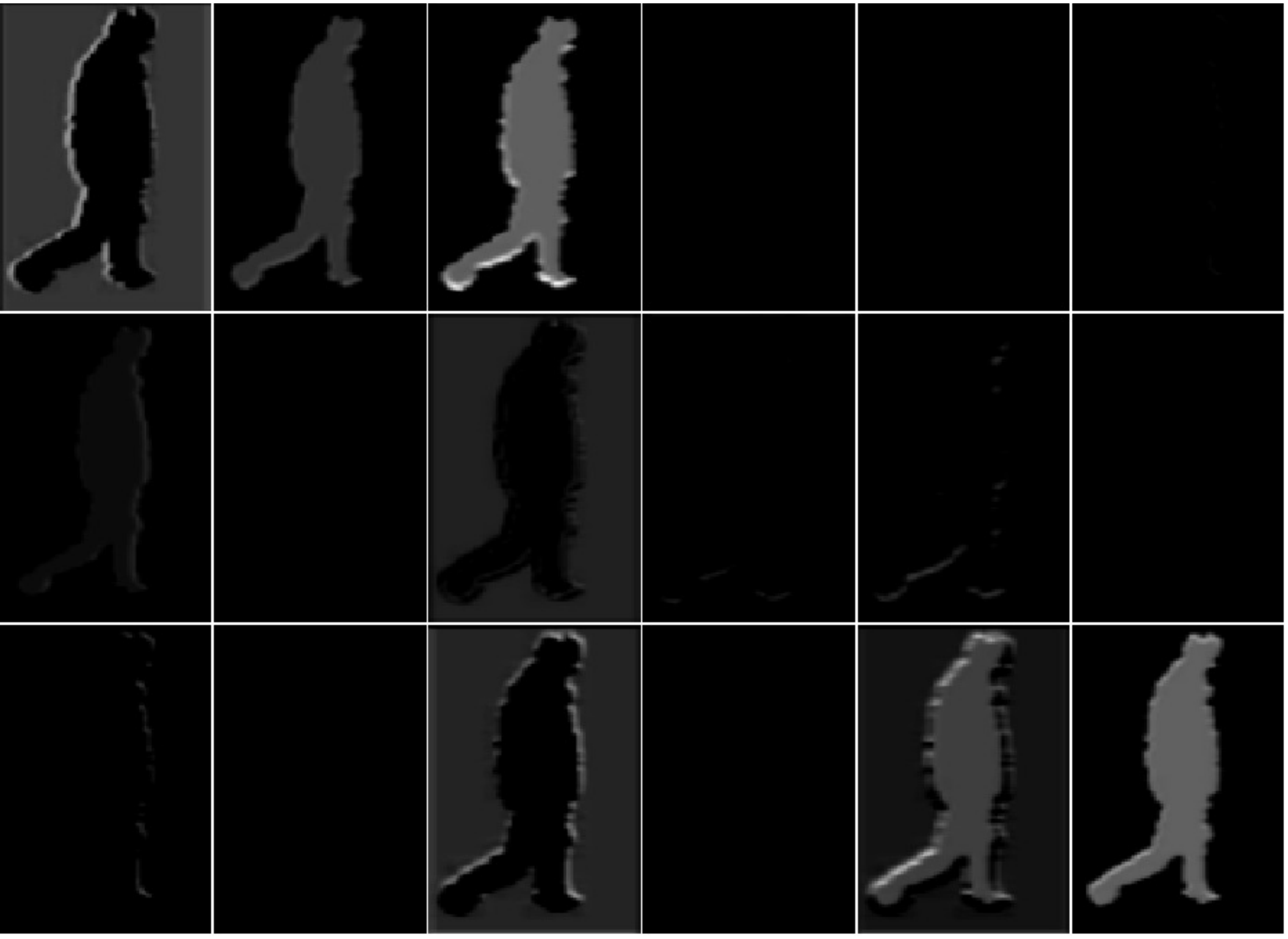}
  \end{minipage}%
  \begin{minipage}{.5\columnwidth}
    \hspace{-0.7mm}
    \includegraphics[width=2.95cm]{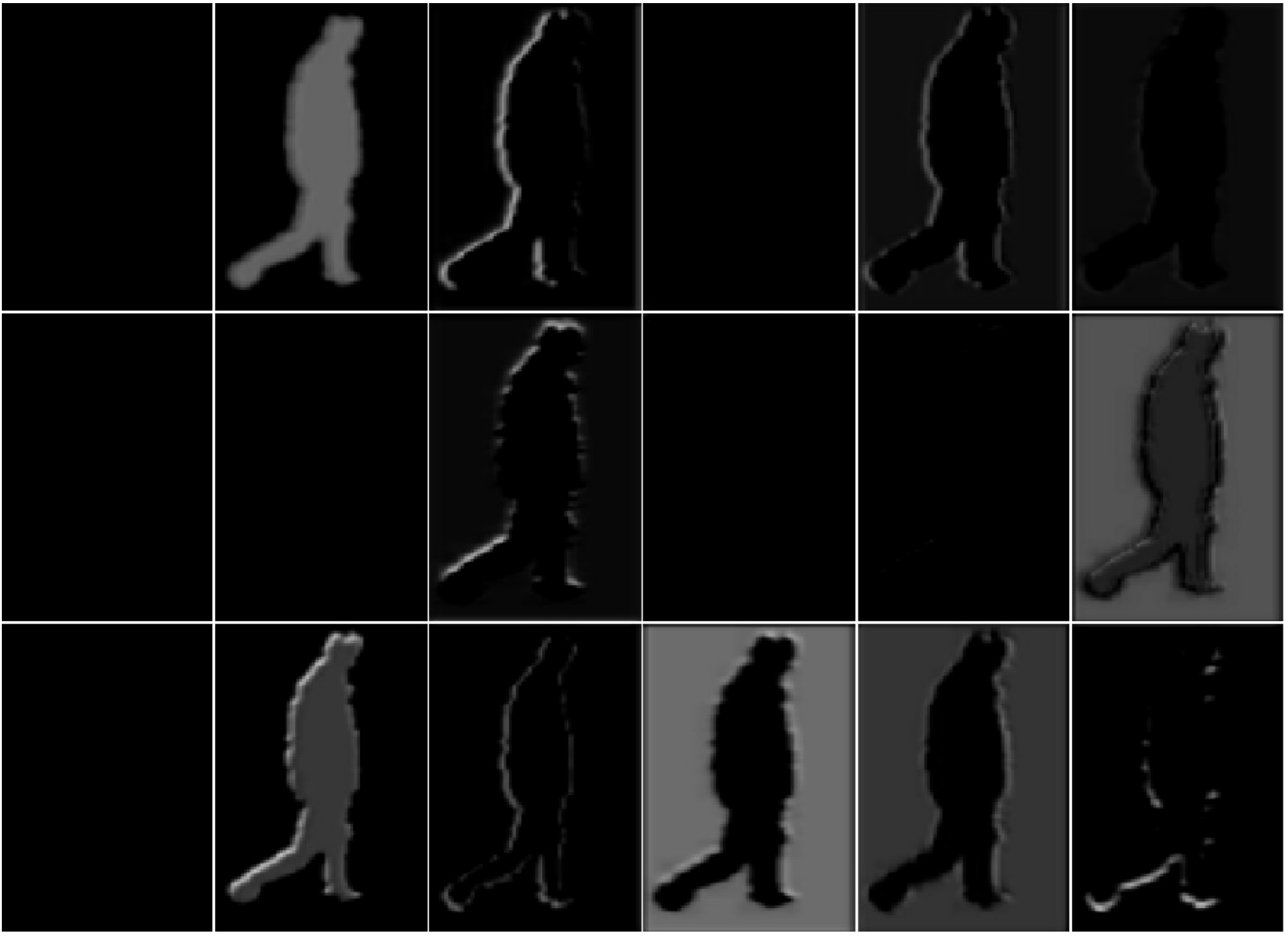}
  \end{minipage}
\vspace{1mm}
\\(b) Person ReID from Depth
 \end{center}
\end{minipage}
\caption{Filter responses from ``conv1" (upper right), ``conv2" (bottom left) and ``conv3" (bottom right) layers for a given frame from 
the TUM GAID data using (a) a framework for person re-identification from RGB~\cite{xiaoLOW16} and (b) the feature embedding $f_{CNN}$ 
of our framework, which is drawn in Fig.~\ref{fig_model} and exclusively utilizes depth data.}
\label{fig_filters}
\end{figure}

\subsection{Related work}
\label{relatedwork}

Existing methods of person re-identification typically focus on designing invariant and discriminant features~\cite{grayT08,farenzenaBPCM10,maSJ12,kviatkovskyAR13,zhaoOW14,lisantiMBB14,yangYYLYL14,castroJC14,liaoHZL15}, 
which can enable identification despite nuisance factors such as scale, location, partial occlusion and changing lighting conditions. 
In an effort to improve their robustness, the current trend is to deploy higher-dimensional descriptors~\cite{liaoHZL15,lisantiMB15} and 
deep convolutional architectures~\cite{liZXW14,yiLL14,ahmedJM15,xiaoLOW16,wangZLZZ16,Zhou_2017_CVPR_point,suZXGT16,Lin_2017_CVPR,Zheng_2017_ICCV,Qian_2017_ICCV,Chung_2017_ICCV,Xiao_2017_CVPR}.

In spite of the ongoing quest for effective representations, it is still challenging to deal with very large variations 
such as ultra wide-baseline matching and dramatic changes in illumination and resolution, especially with limited training data. 
As such, there is vast literature in learning discriminative distance metrics~\cite{koestingerHWRB12,liCLHCS13,zhengGX13,maYT14,mignonJ12,taoJWYL13,liaoHZL15,paisitkriangkraiSH15,martinelDMC16,dingLWC15,Liu_2017_ICCV,Zhou_2017_ICCV,Yu_2017_ICCV,Bak_2017_CVPR} 
and discriminant subspaces~\cite{pedagadiOVB13,lisantiMB15,Zhong_2017_CVPR,xiongGCS14,liaoHZL15,lisantiMBB14,prosserZGX10,zhangXG16,Chen_2017_CVPR}. 
Other approaches handle the problem of pose variability by explicitly accounting for spatial constraints of the human body parts~\cite{chenYCZ16,Zhao_2017_ICCV,Li_2017_CVPR,Zhao_2017_CVPR} 
or by predicting the pose from video~\cite{choY16,Su_2017_ICCV}. 

However, a key challenge to tackle within both distance learning and deep learning pipelines in practical applications is the \emph{small sample size} 
problem \cite{chenLKLY00,zhangXG16}. This issue is exacerbated by the lack of large-scale person re-identification datasets. 
Some new ones have been released recently, such as CUHK03 \cite{liZXW14} and MARS \cite{zhengBSWSWT16}, a video 
extension of the Market-1501 dataset~\cite{zhengSTWWT15}. However, their training sets are in the order of $20,000$ positive samples, 
i.e. two orders of magnitude smaller than Imagenet~\cite{russakovsky15}, which has been successfully used 
for object recognition~\cite{krizhevskySH12,simonyanZ15,szegedyLJSRAEVR15}.

The small sample size problem is especially acute in person re-identification from temporal sequences 
\cite{haqueAF16,yanNSMYY16,castroJGB,mclaughlinMM16,Zhou_2017_CVPR}, as the feature dimensionality increases linearly in the number of frames 
that are accumulated compared to the single-shot representations. On the other hand, explicitly modeling temporal dynamics and 
using multiple frames help algorithms to deal with noisy measurements, occlusions, adverse poses and lighting.

Regularization techniques, such as Batch Normalization~\cite{ioffeS15} and Dropout~\cite{hintonSKSS12}, 
help learning models with larger generalization capability. Xiao et al.~\cite{xiaoLOW16} achieved top accuracy 
on several benchmarks by leveraging on their proposed ``domain-guided dropout" principle. 
After their model is trained on a union of datasets, it is further enhanced on individual datasets by adaptively setting 
the dropout rate for each neuron as a function of its activation rate in the training data.

Haque et al.~\cite{haqueAF16} designed a \emph{glimpse} layer and used a 4D convolutional autoencoder in order to compress 
the 4D spatiotemporal input video representation, 
while the next spatial location (glimpse) is inferred within a recurrent attention framework using reinforcement learning~\cite{mnihHGK14}. 
However, for small patches (at the glimpse location), 
the model loses sight of the overall body shape, while for large patches, it loses the depth resolution. 
Achieving a good \emph{trade-off} between visibility and resolution 
within the objective of compressing the input space to tractable levels is hard with limited data. 
Our algorithm has several key differences from this work. First, observing that there are large amount of RGB data available 
for training frame-level person ReID models, we transfer parameters from pre-trained RGB models 
with an improved transfer scheme. Second, since the input to our frame-level model is the entire body region, 
we do not have any visibility constraints at a cost of resolution. Third, in order to better utilize the temporal information from video, 
we propose a novel reinforced temporal attention unit on top of the frame-level features which is \emph{guided} by the task 
in order to predict the weights of individual frames into the final prediction.

Our method for transferring a RGB Person ReID model to the depth domain is based on the key observation that the model parameters at
the bottom layers of a deep convolutional neural network can be directly shared between RGB and depth data while the remaining upper layers 
need to be fine-tuned. At first glance, our observation is inconsistent with what was reported in the RGB-D object recognition approach 
by Song et al.~\cite{song2017depth}. They reported that the bottom layers cannot be shared between RGB and depth models 
and it is better to retrain them from scratch. Our conjecture is that this behavior is in part specific to the HHA depth encoding~\cite{gupta2014learning}, 
which is not used in our representation.

Some recent works in natural language processing~\cite{luong2015effective,chan2015listen} 
explore temporal attention in order to keep track of long-range structural dependencies. Yao et al.~\cite{yao2015describing} in video captioning 
use a soft attention gate inside their Long Short-term memory decoder, so that they estimate the relevance of current features in the input video 
given all the previously generated words. One key difference of our approach is that our attention unit is exclusively dependent on the frame-level 
feature embedding, but not on the hidden state, which likely makes it less prone to error drifting. Additionally, our temporal attention is 
not differentiable so we resort to reinforcement learning techniques~\cite{williams92} for binary outcome. Being inspired by the work of Likas~\cite{likas99} 
in online clustering and Kontoravdis et al.~\cite{kontoravdis95} in exploration of binary domains, we model the weight of each frame prediction as 
a \emph{Bernoulli-sigmoid} unit. We review our model in detail in Sec.~\ref{model}.

Depth-based methods that use measurements from 3D skeleton data have emerged in order to infer anthropometric and 
human gait criteria \cite{munsellTQW12,mogelmoseMN13,albiolOM12,anderssonDA,duboisC14}. In an effort to leverage the full power of depth data, 
recent methods use 3D point clouds to estimate motion trajectories and the length of specific body parts~\cite{ioannidisTDAM07,zhaoLLP}. 
It is worthwhile to point out that skeleton information is not always available. For example, the skeleton tracking in Kinect SDK can be ineffective 
when a person is in side view or the legs are not visible.

On top of the above-mentioned challenges, RGB-based methods are challenged in scenarios with significant lighting 
changes and when the individuals change clothes. These factors can have a big impact on the effectiveness of a system that, for 
instance, is meant to track people across different areas of a building over several days where different areas of a building may have 
drastically different lighting conditions, the cameras may differ in color balance, and a person may wear clothes of different patterns. 
This is our \emph{key motivation} for using depth silhouettes in our scenario, as they are insensitive to these factors.


Our contributions can be summarized as follows:\\
\indent i) We propose novel reinforced temporal attention on top of the frame-level features to better leverage the temporal information 
from video sequences by learning to adaptively weight the predictions of individual frames based on a task-based reward.
In Sec.~\ref{model} we define the model, its end-to-end training is described in Sec.~\ref{training}, and comparisons with baselines 
are shown in Sec.~\ref{results}.\\
\indent ii) We tackle the data scarcity problem in depth-based person re-identification by leveraging the large amount of RGB data to obtain stronger 
frame-level features. Our \emph{split-rate} RGB-to-depth transfer scheme is drawn in Fig.~\ref{plot_transfer_method}. 
We show in Fig.~\ref{plot_transfer_results} that our method outperforms a popular fine-tuning method by more effectively utilizing pre-trained models from RGB data.\\
\indent iii) Extensive experiments in Sec.~\ref{results} not only show the superiority of our method compared to the state of the art 
in depth-based person re-identification from video, but also tackle a challenging application scenario where the persons wear clothes 
that were unseen during training. In Table~\ref{table_tumgaid} we demonstrate the robustness of our method compared to its RGB-based counterpart 
and the mutual gains when jointly using the person's head information.

\section{Our Method}
\label{method}

\subsection{Input Representation}
\label{representation}

The input for our system is raw depth measurements from the Kinect V$2$~\cite{shottonSKFFBCM13}. The input data are depth images 
${\bf D}\in \mathbb{Z}^{512\times424}$, where each pixel $D[i,j], i\in [1,512], j\in [1,424]$, contains the Cartesian distance, 
in millimeters, from the image plane to the nearest object at the particular coordinate $(i,j)$. In ``default range" setting, the intervals $[0, 0.4m)$ and 
$(8.0m, \infty)$ are classified as unknown measurements, $[0.4, 0.8) [m]$ as ``too near", $(4.0, 8.0] [m]$ as ``too far" 
and $[0.8, 4.0] [m]$ as ``normal" values. When skeleton tracking is effective, the \emph{body index} ${\bf B}\in \mathbb{Z}^{512\times424}$ 
is provided by the Kinect SDK, where $0$ corresponds to background and a positive integer $i$ for each pixel belonging to the person $i$. 

\begin{figure}[t]
\begin{center}
$\vimage{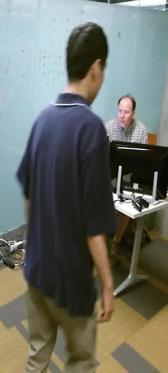}\hspace{12mm}
\vimage{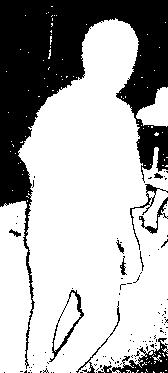}\vpointer
\vimage{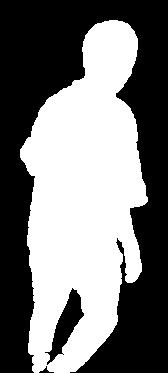}$
\end{center}
\caption{The cropped color image (left), the grayscale depth representation ${\bf D_p^g}$ (center) and the result after background subtraction (right) 
using the body index information ${\bf B_p}$ from skeleton tracking.}
\label{fig_representations}
\end{figure}

After extracting the person region ${\bf D_p} \subset {\bf D}$, the measurements within the ``normal" region are normalized in the range 
$[1, 256]$, while the values from ``too far" and ``unknown" range are set as $256$, and values within the ``too near" range as $1$. 
In practice, in order to avoid a concentration of the values near $256$, whereas other values, say on the floor in front of the subject, 
span the remaining range, we introduce an offset $t_o=56$ and normalize in $[1, 256-t_o]$. This results in the ``grayscale" person 
representation ${\bf D_p^g}$. When the body index is available, we deploy ${\bf B_p} \subset {\bf B}$ as mask on the depth region ${\bf D_p}$ 
in order to achieve background subtraction before applying range normalization (see Fig.~\ref{fig_representations}). 

\subsection{Model Structure}
\label{model}

\begin{figure}[t]
\begin{center}
  \includegraphics[width=7.5cm]{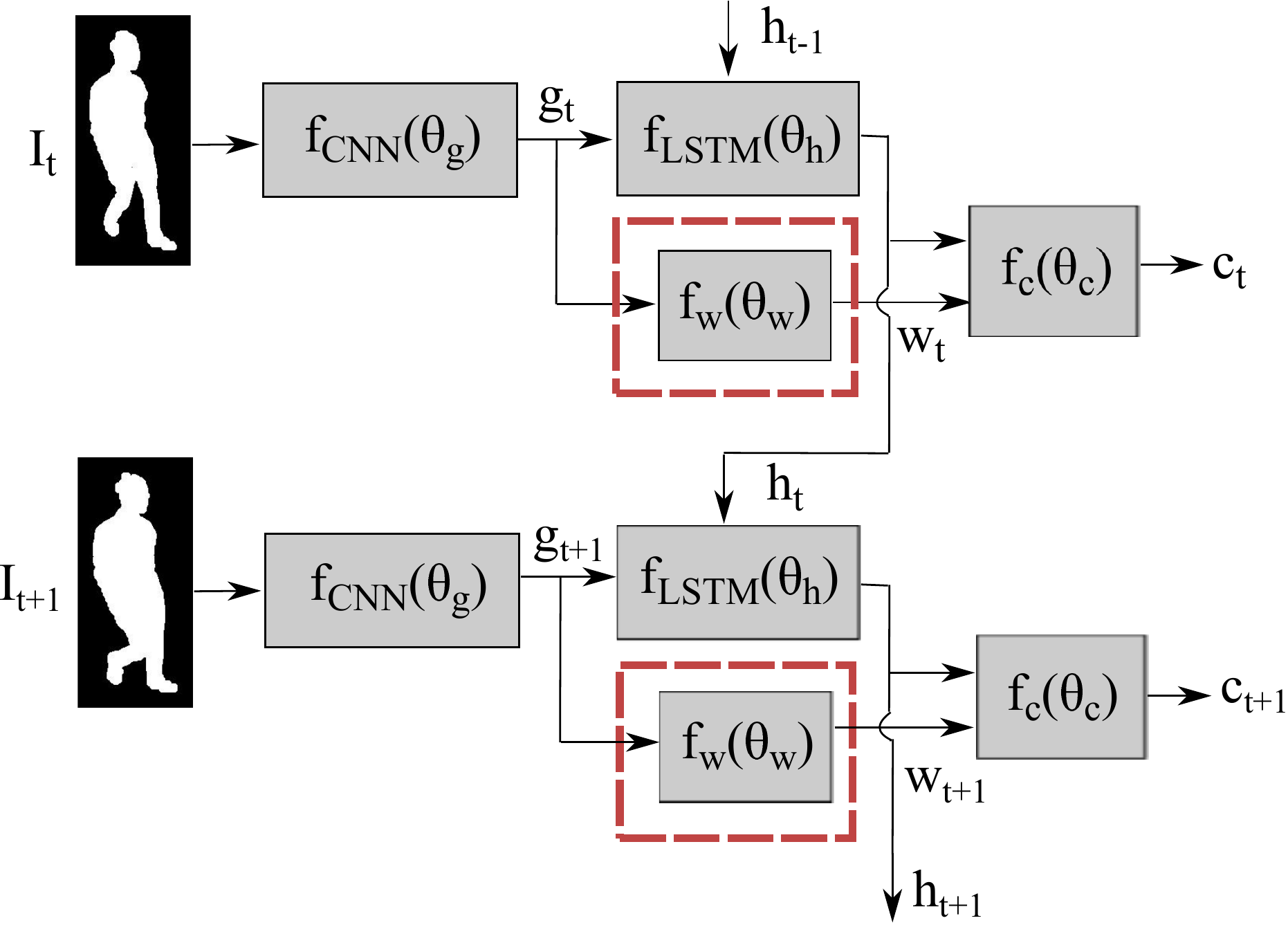}
\end{center}
\caption{Our model architecture consists of a frame-level feature embedding $f_{CNN}$, which provides input to both a recurrent layer $f_{LSTM}$ 
and the Reinforced Temporal Attention (RTA) unit $f_w$ (highlighted in red). The classifier is attached to the hidden state $h_t$ and 
its video prediction is the weighted sum of single-frame predictions, where the weights $w_t$ for each frame $t$ are predicted by the RTA unit.}
\label{fig_model}
\end{figure}

The problem is formulated as \textit{sequential decision process} of an agent that performs human recognition 
from a partially observed environment via video sequences. At each time step, the agent observes 
the environment via depth camera, calculates a feature vector based on a deep Convolutional Neural Network (CNN) 
and actively infers the importance of the current frame for the re-identification task using novel Reinforced Temporal Attention (RTA). 
On top of the CNN features, a Long Short-Term Memory (LSTM) unit models short-range temporal dynamics. 
At each time step the agent receives a reward based on the success or failure of its classification task. Its objective is to maximize 
the sum of rewards over time. The agent and its components are detailed next, while the training process is described in Sec.~\ref{training}. 
The model is outlined in Fig.~\ref{fig_model}.

\subsubsection{Agent:}

Formally, the problem setup is a Partially Observable Markov Decision Process (POMDP). The true state of the environment is unknown. 
The agent learns a stochastic policy $\pi((w_t, c_t)|s_{1:t}; \theta)$ with parameters $\theta=\{\theta_g, \theta_w, \theta_h, \theta_c \}$ that, 
at each step $t$, maps the past history $s_{1:t}=I_1, w_1, c_1, \ldots, I_{t-1}, w_{t-1}, c_{t-1}, I_t$ to two distributions over discrete actions: 
the frame weight $w_t$ (sub-policy $\pi_1$) and the class posterior $c_t$ (sub-policy $\pi_2$). The weight $w_t$ is sampled stochastically 
from a binary distribution parameterized by the RTA unit $f_w(g_t;\theta_w)$ at time $t$: $w_t \sim \pi_1(\cdot | f_w(g_t;\theta_w))$. The class posterior 
distribution is conditioned on the classifier module, which is attached to the LSTM output $h_t$: $c_t \sim \pi_2(\cdot | f_c(h_t;\theta_c))$. 
The vector $h_t$ maintains an internal state of the environment as a summary of past observations. 
Note that, for simplicity of notation, the input image at time $t$ is denoted as $I_t$, but the actual input is the person region $D_{p,t}^g$ 
(see Sec.~\ref{representation}).

\subsubsection{Frame-level Feature Embedding $f_{CNN}(\theta_g)$:}
 
Given that there is little depth data but a large amount of RGB data available for person re-identification, 
we would like to leverage the RGB data to train depth models for frame-level feature extraction. 
We discovered that the parameters at the bottom convolutional layers of a deep neural network can be directly 
shared between RGB and depth data (\emph{cf.}~Sec.~\ref{training}) through a simple depth encoding, that is, each pixel with depth D 
is replicated to three channels and encoded as (D,D,D), which corresponds to the three RGB channels. This motivates us to select a pre-trained RGB model.

RGB-based person re-identification has progressed rapidly in recent years \cite{liZXW14,yiLL14,ahmedJM15,xiaoLOW16,wangZLZZ16,suZXGT16}. 
The deep convolutional network introduced by Xiao et al.~\cite{xiaoLOW16} outperformed other approaches on several public datasets. 
Therefore, we decide to adopt their model for frame-level feature extraction. This network
is similar in nature to \emph{GoogleNet}~\cite{szegedyLJSRAEVR15}; it uses batch normalization~\cite{ioffeS15} and includes 
$3\times3$ convolutional layers~\cite{simonyanZ15}, followed by $6$ Inception modules~\cite{szegedyLJSRAEVR15}, and $2$ fully connected layers. 
In order to make this network applicable to our scenario, we introduce two small modifications. First, we replace the top classification layer 
with a $256\times N$ fully connected layer, where $N$ is the number of subjects at the target dataset and its weights are initialized at random 
from a zero-mean Gaussian distribution with standard deviation $0.01$. Second, we add dropout regularization between the fully-connected layers. 
In Sec.~\ref{training} we demonstrate an effective way to transfer the model parameters from RGB to Depth.

\subsubsection{Recurrent Module $f_{LSTM}(\theta_h)$:}

We use the efficient Long Short-Term Memory (LSTM) element units as described in \cite{zarembaS14}, which have been shown 
by Donahue et al.~\cite{donahueHGRVSD15} to be effective in modeling temporal dynamics for video recognition and captioning. 
In specific, assuming that $\sigma()$ is sigmoid, $g[t]$ is the input at time frame $t$, $h[t-1]$ is the previous output of the module 
and $c[t-1]$ is the previous cell, the implementation corresponds to the following updates:
\begin{align}
i[t] &= \sigma(W_{gi}g[t] + W_{hi}h[t-1] + b_i)\\
f[t] &= \sigma(W_{gf}g[t] + W_{hf}h[t-1] + b_f)\\
z[t] &= tanh(W_{gc}g[t] + W_{hc}h[t-1] + b_c)\\
c[t] &= f[t] \odot c[t-1] + i[t] \odot z[t]\\
o[t] &= \sigma(W_{go}g[t] + W_{ho}h[t-1] + b_o)\\
h[t] &= o[t] \odot tanh(c[t])
\end{align}
where $W_{sq}$ is the weight matrix from source $s$ to target $q$ for each gate $q$, $b_q$ are the biases leading into $q$, $i[t]$ is the input gate, 
$f[t]$ is the forget gate, $z[t]$ is the input to the cell, $c[t]$ is the cell, $o[t]$ is the output gate, and $h[t]$ is the output of this module. 
Finally, $x \odot y$ denotes the element-wise product of vectors $x$ and $y$.

\subsubsection{Reinforced Temporal Attention $f_w(\theta_w)$:}

At each time step $t$ the RTA unit infers the \emph{importance} $w_t$ of the image frame $I_t$, as the latter is represented 
by the feature encoding $g_t$. This module consists of a linear layer which maps the $256\times1$ vector $g_t$ to one scalar, 
followed by Sigmoid non-linearity which squashes real-valued inputs to a $[0,1]$ range. Next, the output $w_t$ is defined 
by a Bernoulli random variable with probability mass function:
\begin{equation}
    f(w_t; f_w(g_t;\theta_w)) =
    \begin{cases}
      f_w(g_t;\theta_w), &  w_t=1 \\
      1-f_w(g_t;\theta_w), &  w_t=0
    \end{cases}
\end{equation}

\noindent The Bernoulli parameter is conditioned on the Sigmoid output $f_w(g_t;\theta_w)$, shaping a Bernoulli-Sigmoid unit~\cite{williams92}. 
During training, the output $w_t$ is sampled \textit{stochastically} to be a binary value in $\{0,1\}$. 
During evaluation, instead of sampling from the distribution, the output is deterministically decided to be equal to the Bernoulli parameter 
and, therefore, $w_t=f_w(g_t;\theta_w)$.


\subsubsection{Classifier $f_c(\theta_c)$ and Reward:}

The classifier consists of a sequence of a rectified linear unit, dropout with rate $r=0.4$, a fully connected layer and Softmax. 
The parametric layer maps the $256\times1$ hidden vector $h_t$ to the $N\times1$ class posterior vector $c_t$, which has length 
equal to the number of classes $N$. The multi-shot prediction with RTA attention is the weighted sum of frame-level predictions $c_t$, 
as they are weighted by the normalized, RTA weights $w_t' = \frac{f_w(g_t;\theta_w)}{\sum_{t=1}^T f_w(g_t;\theta_w)} $.

The Bernoulli-Sigmoid unit is stochastic during training and therefore we resort to the REINFORCE algorithm in order to obtain 
the gradient for the backward pass. We describe the details of the training process in Sec.~\ref{training}, 
but here we define the required reward function. A straightforward definition is:
\begin{equation}
r_t = \mathcal{I} (arg\,max (c_t) = g_t)
\label{eq_reward}
\end{equation}
where $r_t$ is the raw reward, $\mathcal{I}$ is the indicator function and $g_t$ is the ground-truth class for frame $t$. Thus, at each time step $t$, 
the agent receives a reward $r_t$, which equals to $1$ when the frame is correctly classified and $0$ otherwise.

\subsection{Model Training}
\label{training}

In our experiments we first pre-train the parameters of the frame-level feature embedding, and afterwards we attach LSTM, RTA and the new classifier 
in order to train the whole model (\emph{cf.} Fig.~\ref{fig_model}). At the second step the weights of the embedding are frozen while the added layers are initialized at random. 
We adopt this modular training so that we provide both single-shot and multi-shot evaluation, but the entire architecture can well be trained end to end 
from scratch if processing video sequences is the sole objective. Next, we first describe our transfer learning for the frame-level embedding 
and following the hybrid supervised training algorithm for the recursive model with temporal attention.

\subsubsection{Split-rate Transfer Learning for Feature Embedding $f_{CNN}(\theta_g)$:}

In order to leverage on vast RGB data, our approach relies on transferring parameters $\theta_g$ from a RGB pre-trained model for initialization. 
As it is unclear whether and which subset of RGB parameters is beneficial for depth embedding, we first gain insight from work by Yosinski et al.~\cite{yosinskiCBL14} 
in CNN feature transferability. They showed that between two almost equal-sized splits from Imagenet~\cite{russakovsky15}, the most effective 
model adaptation is to transfer and slowly fine-tune the weights of the bottom convolutional layers, while re-training the top layers. 
Other works that tackle model transfer from a large to a small-sized dataset (\emph{e.g.}~\cite{karayev2013recognizing}) copy and slowly fine-tune 
the weights of the whole hierarchy except for the classifier which is re-trained using a higher learning rate. 

\begin{figure}[t]
\begin{center}
  \includegraphics[width=9.5cm]{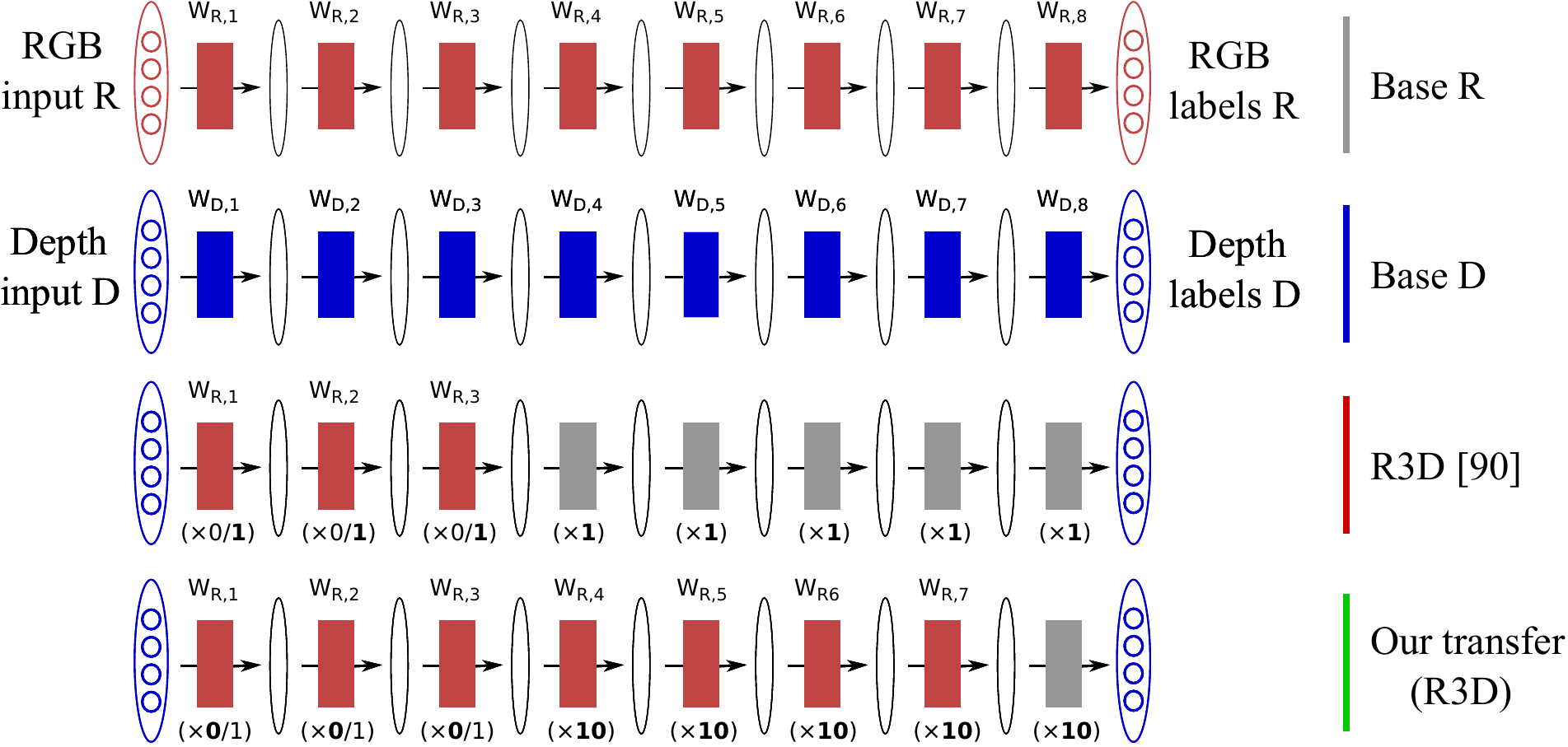}
\end{center}
\caption{Our split-rate RGB-to-Depth transfer compared with Yosinski et al.~\cite{yosinskiCBL14}. At the top, the two models are trained from scratch 
with RGB and Depth data. Next we show the ``R3D" instances (i.e. the bottom $3$ layers' weights from RGB remain frozen or slowly changing) for both methods, 
following the notation of \cite{yosinskiCBL14}. The color of each layer refers to the initialization and the number below is the relative learning rate 
(the best performing one in bold). The key differences are summarized in the text.}
\label{plot_transfer_method}
\end{figure}

Inspired by both approaches, we investigate the model transferability between RGB and depth. 
Our method has three differences compared to~\cite{yosinskiCBL14}. 
First, we found that even though RGB and depth are quite different modalities (\emph{cf.}~Fig.~\ref{fig_filters}), 
the bottom layers of the RGB models can be shared with the depth data (without fine-tuning). 
Second, fine-tuning parameters transferred from RGB works better than training from scratch for the top layers. 
Third, using slower (or zero) learning rate for the bottom layers and higher for the top layers is more effective than using uniform rate 
across the hierarchy. Thus, we term our method as \emph{split-rate} transfer. 
This first and third remarks also consist key differences with~\cite{karayev2013recognizing}, as firstly they fine-tune all layers and secondly 
they deploy higher learning rate only for the classifier. Our approach is visualized in Fig.~\ref{plot_transfer_method} and ablation studies 
are shown in Sec.~\ref{experiments_transfer} and Fig.~\ref{plot_transfer_results}, which support the above-mentioned observations.


\subsubsection{Hybrid Learning for CNN-LSTM and Reinforced Temporal Attention:}

The parameters $\{\theta_g, \theta_h, \theta_c \}$ of CNN-LSTM are learned by minimizing the classification loss that is attached 
on the LSTM unit via backpropagation backward through the whole network. We minimize the cross-entropy loss as customary in recognition tasks, 
such as face identification~\cite{sunWT14}. Thus, the objective is to maximize the conditional probability of the true label given the observations, 
i.e. we maximize $\log \pi_2 (c_t^* | s_{1:t}; \theta_g, \theta_h, \theta_c)$, where $c_t^*$ is the true class at step $t$.

The parameters $\{\theta_g, \theta_w\}$ of CNN and RTA are learned so that the agent maximizes its total reward $R=\sum_{t=1}^T r_t$, 
where $r_t$ has defined in Eq.~\ref{eq_reward}. This involves calculating the expectation $J(\theta_g, \theta_w)=\mathbb{E}_{p(s_{1:T}; \theta_g, \theta_w)}[R]$ 
over the distribution of all possible sequences $p(s_{1:T}; \theta_g, \theta_w)$, which is intractable. Thus, a sample approximation, 
known as the REINFORCE rule~\cite{williams92}, can be applied on the Bernoulli-Sigmoid unit~\cite{likas99,kontoravdis95}, which models 
the sub-policy $\pi_1(w_t | f_w(g_t;\theta_w))$. Given probability mass function $\log \pi_1(w_t; p_t)=w_t\log p_t+(1-w_t)\log(1-p_t)$ 
with Bernoulli parameter $p_t=f_w(g_t;\theta_w)$, the gradient approximation is:
\begin{align}
\nabla_{\theta_g, \theta_w}J &= \sum_{t=1}^T \mathop{\mathbb{E}}\nolimits_{p(s_{1:T}; \theta_g, \theta_w)} [\nabla_{\theta_g, \theta_w}\log\pi_1(w_t | s_{1:t};\theta_g, \theta_w)(R_t-b_t)]\\
                 & \approx \frac{1}{M} \sum_{i=1}^M \sum_{t=1}^T \frac{w_t^i-p_t^i}{p_t^i(1-p_t^i)} (R_t^i-b_t)
\end{align}
where sequences $i$, $i\in \{1,\ldots,M\}$, are obtained while running the agent for $M$ episodes and $R_t^i=\sum_{\tau=1}^t r_{\tau}^i$ is 
the cumulative reward at episode $i$ acquired after collecting the sample $w_t^i$. The gradient estimate is biased by a baseline reward $b_t$ 
in order to achieve lower variance. Similarly to \cite{mnihHGK14,haqueAF16}, we set $b_t = \mathbb{E}_{\pi}[R_t]$, as the mean square error 
between $R_t^i$ and $b_t$ is also minimized by backpropagation.

At each step $t$, the agent makes a prediction $w_t$ and the reward signal $R_t^i$ evaluates the effectiveness of the agent for the classification task. 
The REINFORCE update increases the log-probability of an action that results in higher than the expected accumulated reward 
(i.e. by increasing the Bernoulli parameter $f_w(g_t;\theta_w)$). Otherwise, the log-probability decreases for sequence of frames that lead to low reward. 
All in all, the agent jointly optimizes the accumulated reward and the classification loss, which constitute a \textit{hybrid} supervised objective.


\section{Experiments}
\label{experiments}

\subsection{Depth-based Datasets}
\label{datasets}

\paragraph{DPI-T (Depth-based Person Identification from Top).}
Being recently introduced by Haque et al.~\cite{haqueAF16}, it contains $12$ persons appearing in a total of $25$ sequences 
across many days and wearing $5$ different sets of clothes on average. 
Unlike most publicly available datasets, the subjects appear from the top, which is a common scenario in automated video surveillance. 
The individuals are captured in daily life situations where they hold objects such as handbags, laptops and coffee.

\paragraph{BIWI.} In order to explore sequences with varying human pose and scale, 
we use BIWI~\cite{munaroBFGM14}, where $50$ individuals appear in a living room. $28$ of them are re-recorded 
in a different room with new clothes and walking patterns. We use the full training set, while for testing we use the \emph{Walking} set. 
From both sets we remove the frames with no person or a person heavily occluded from the image boundaries or too far from the sensor, 
as they provide no skeleton information.

\paragraph{IIT PAVIS.} To evaluate our method when shorter video sequences are available, we use IIT PAVIS~\cite{barbosaCBBM12}. 
This dataset includes $79$ persons that are recorded in 5-frame walking sequences twice. We use \emph{Walking1} and \emph{Walking2} 
sequences as the training and testing set, respectively.

\paragraph{TUM-GAID.} To evaluate on a large pool of identities, we use the \emph{TUM-GAID} database~\cite{hofmannGBSR14}, 
which contains RGB and depth video for $305$ people in three variations. A subset of $32$ people is recorded a second time 
after three months with different clothes, which makes it ideal for our application scenario in Sec.~\ref{tumgaid}. In our experiments 
we use the ``normal" sequences (n) from each recording.

\subsection{Evaluation Metrics}
\label{metrics}

\emph{Top-k accuracy} equals the percentage of test images or sequences for which the ground-truth label is contained within 
the first $k$ model predictions. Plotting the top-k accuracy as a function of $k$ gives the \emph{Cumulative Matching Curve} (CMC). 
Integrating the area under the CMC curve and normalizing over the number of IDs produces the normalized \emph{Area Under the Curve} (nAUC). 

In single-shot mode the model consists only of the $f_{CNN}$ branch with an attached classifier (see Fig.~\ref{fig_model}). In multi-shot mode, 
where the model processes sequences, we evaluate our CNN-LSTM model with (or without) RTA attention.


\subsection{Experimental Setting}
\label{settings}

The feature embedding $f_{CNN}$ is trained in Caffe~\cite{jiaSDKLGGD14}. Consistent with~\cite{xiaoLOW16}, the input depth images 
are resized to be $144\times56$. SGD mini-batches of $50$ images are used for training and testing. 
Momentum $\mu=0.5$ yielded more stable training. The momentum effectively multiplies the size of the updates 
by a factor of $\frac{1}{1-\mu}$ after several iterations, so lower values result in smaller updates. The weight decay is set to $2*10^{-4}$, 
as it is common in Inception architecture~\cite{szegedyLJSRAEVR15}. We deploy modest base learning rate $\gamma_0=3\times10^{-4}$. 
The learning rate is reduced by a factor of $10$ throughout training every time the loss reaches a ``plateau".

The whole model with the LSTM and RTA layers in Fig.~\ref{fig_model} is implemented in Torch/Lua~\cite{collobertKF11}. We implemented 
customized Caffe-to-Torch conversion scripts for the pre-trained embedding, as the architecture is not standard. For end-to-end training, 
we use momentum $\mu=0.9$, batch size $50$ and learning rate that linearly decreases from $0.01$ to $0.0001$ in $200$ epochs 
up to $250$ epochs maximum duration. The LSTM history consists of $\rho=3$ frames.

\begin{figure}[t]
\begin{minipage}{.5\linewidth}
\begin{center}
  \includegraphics[height=4.6cm]{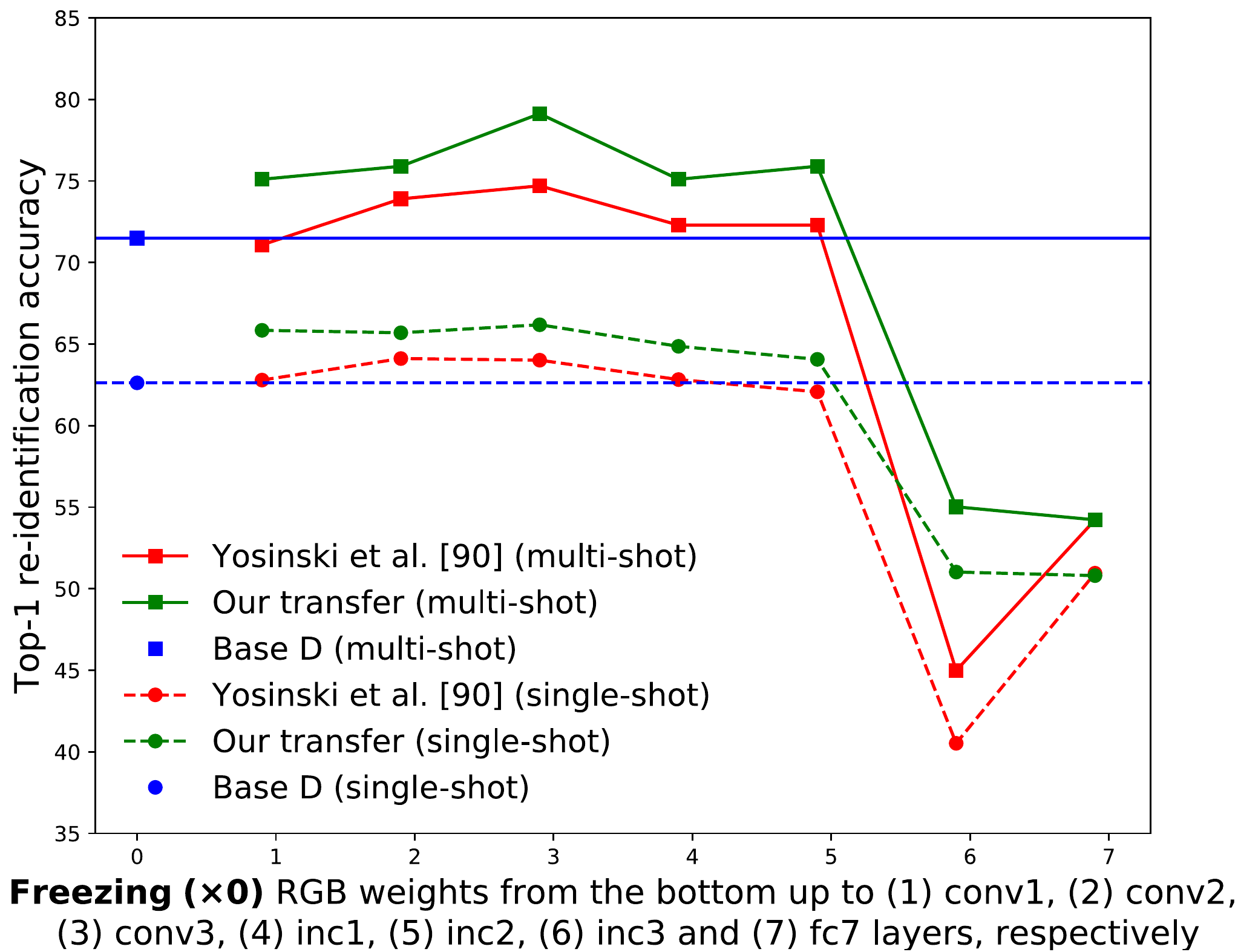}
\end{center}
\end{minipage}%
\hspace{1mm}
\begin{minipage}{.5\linewidth}
\begin{center}
  \includegraphics[height=4.6cm]{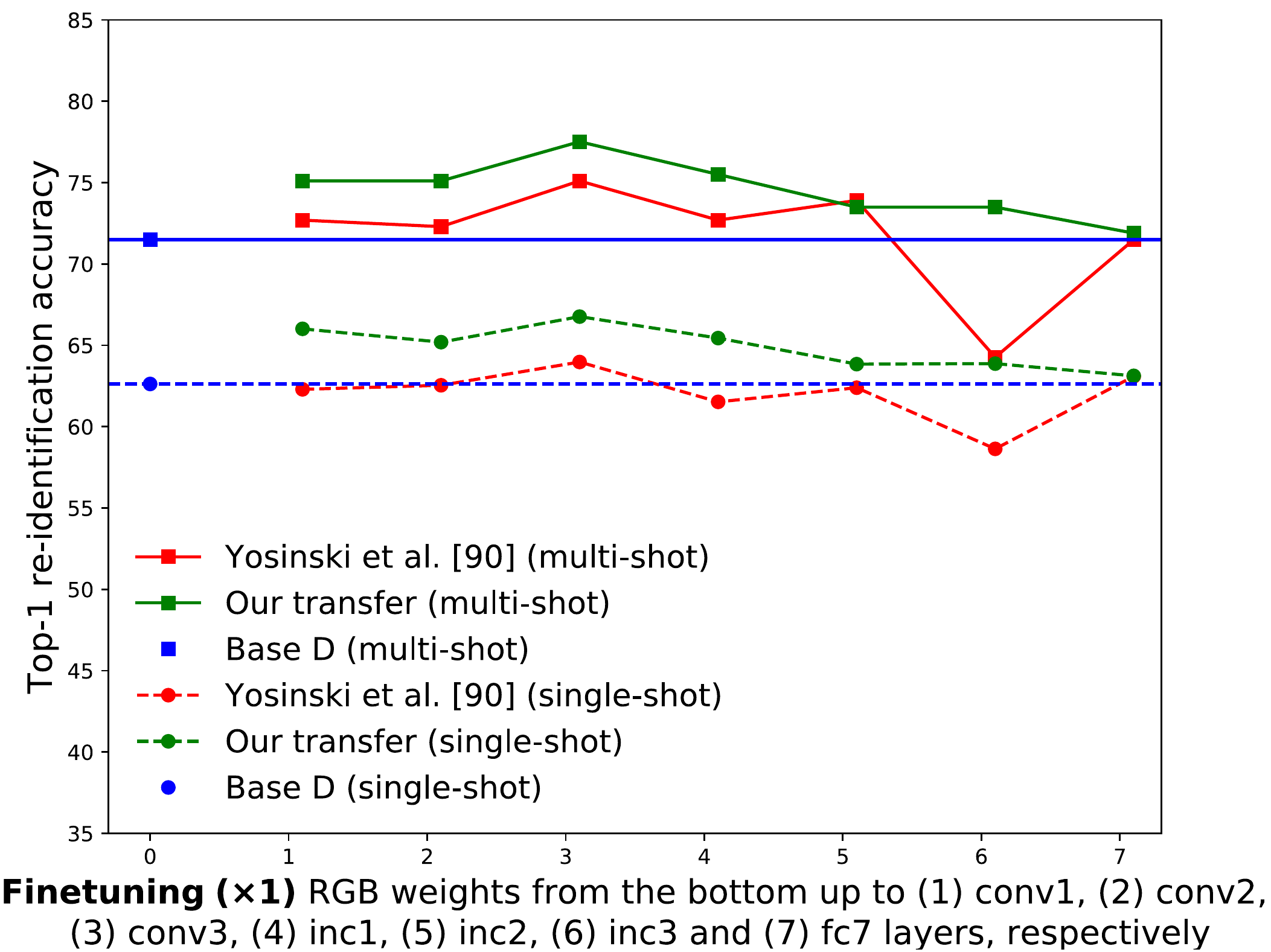}
\end{center}
\end{minipage}
\caption{Comparison of our RGB-to-Depth transfer with Yosinski et al.~\cite{yosinskiCBL14} in terms of top-1 accuracy on DPI-T. In this ablation study 
the x axis represents the number of layers whose weights are frozen (left) or fine-tuned (right) starting from the bottom.}
\label{plot_transfer_results}
\end{figure}

\subsection{Evaluation of the Split-Rate RGB-to-Depth Transfer}
\label{experiments_transfer}

In Fig.~\ref{plot_transfer_results} we show results of our split-rate RGB-to-Depth transfer (which is described in Sec.~\ref{training}) 
compared to~\cite{yosinskiCBL14}. 
We show the top-1 re-identification accuracy on DPI-T when the bottom CNN layers are frozen (left) and slowly fine-tuned (right). 
The top layers are transferred from RGB and rapidly fine-tuned in our approach, while they were re-trained in~\cite{yosinskiCBL14}.
Given that the CNN architecture has $7$ main layers before the classifier, the $x$ axis is the number of layers 
that are frozen or fine-tuned counting from the bottom. 

Evidently, transferring and freezing the three bottom layers, while rapidly fine-tuning the subsequent ``inception" and fully-connected layers, 
brings in the best performance on DPI-T. Attempting to freeze too many layers leads to performance drop for both approaches, which can 
been attributed to feature \emph{specificity}. Slowly fine-tuning the bottom layers helps to alleviate \emph{fragile co-adaptation}, as it was pointed out 
by Yosinski et al.~\cite{yosinskiCBL14}, and improves generalization, especially while moving towards the right of the x axis. Overall, 
our approach is more accurate in our setting across the x axis for both treatments.

\subsection{Evaluation of the End-to-End Framework}
\label{results}

\begin{table*}[t]
\begin{center}
\caption{Single-shot and multi-shot person re-identification performance on the test set of DPI-T, BIWI and IIT PAVIS. Dashes indicate that no published result is available}
\label{table_comparisons}
\begin{tabular}{c | c | c | c | c}
\hline \multirow{2}{*}{Mode} & \multirow{2}{*}{Method} & \multicolumn{3}{c}{Top-1 Accuracy ($\%$)} \\
\cline{3-5} &  & DPI-T & BIWI & IIT PAVIS \\ \hline \hline
                                           & Random                                                   & 8.3  & 2.0 & 1.3  \\ \hline
\multirow{4}{*}{Single-shot} &  Skeleton (NN)~\cite{munaroBFGM14}     & --  & 21.1  & 28.6  \\ \cline{2-5}
                                            &  Skeleton (SVM)~\cite{munaro2014one}   & --  & 13.8  & 35.7  \\ \cline{2-5}
                                            &  3D RAM~\cite{haqueAF16}                      & 47.5 & {\bf 30.1} & 41.3 \\ \cline{2-5}
                                            &  Our method (CNN)                                 & {\bf 66.8} & 25.4 & {\bf 43.0}  \\ \hline
\multirow{8}{*}{Multi-shot}   &   Skeleton (NN)~\cite{munaroBFGM14}     & --  & 39.3 & --  \\ \cline{2-5}
                                           &  Skeleton (SVM)~\cite{munaro2014one}   & --  & 17.9  & --  \\ \cline{2-5}
                                           &  Energy Volume~\cite{sivapalan2011gait}   & 14.2  & 25.7  & 18.9   \\ \cline{2-5}
                      &  3D CNN+Avg Pooling~\cite{boureau2010theoretical}   & 28.4  & 27.8  & 27.5  \\ \cline{2-5}
                                           &  4D RAM~\cite{haqueAF16}                      & 55.6 & 45.3 & 43.0  \\ \cline{2-5}
                                           &  Our method (CNN-LSTM+Avg Pooling )  & 75.5 & 45.7 & 50.1  \\ \cline{2-5}
               &  Our method with attention from~\cite{yao2015describing}    & 75.9 & 46.4 & 50.6  \\ \cline{2-5}
                    &  Our method with RTA attention                                        & {\bf 76.3} & {\bf 50.0}  & {\bf 52.4} \\ \hline
\end{tabular}
\end{center}
\end{table*}

In Table~\ref{table_comparisons} we compare our framework with depth-based baseline algorithms. First, we show 
the performance of guessing uniformly at random. Next, we report results from \cite{barbosaCBBM12,munaro2014one}, 
who use hand-crafted features based on biometrics, such as distances between skeleton joints. A 3D CNN with average pooling 
over time~\cite{boureau2010theoretical} and the gait energy volume~\cite{sivapalan2011gait} are evaluated in multi-shot mode. 
Finally, we provide the comparisons with 3D and 4D RAM models~\cite{haqueAF16}.

In order to evaluate our model in multi-shot mode without temporal attention, we simply average the output of the classifier attached 
on the CNN-LSTM output across the sequence (\emph{cf.} Fig.~\ref{fig_model}). In the last two rows we show results that leverage temporal attention. 
We compare our RTA attention with the \emph{soft} attention in~\cite{yao2015describing}, which is a function of both the hidden state $h_t$ 
and the embedding $g_t$, whose projections are added and passed through a $tanh$ non-linearity. 

We observe that methods that learn end-to-end re-identification features perform significantly better 
than the ones that rely on hand-crafted biometrics on all datasets. Our algorithm is the top performer in multi-shot mode, 
as our RTA unit effectively learns to re-weight the most effective frames based on classification-specific reward. 
The split-rate RGB-to-Depth transfer enables our method to leverage on RGB data effectively and provides 
discriminative depth-based ReID features. This is especially reflected by the single-shot accuracy on DPI-T, where 
we report $19.3\%$ better top-1 accuracy compared to 3D RAM. However, it is worth noting that 3D RAM performs better on BIWI. 
Our conjecture is that the spatial attention mechanism is important in datasets with significant variation in human pose and partial body occlusions. 
On the other hand, the spatial attention is evidently less critical on DPI-T, which contains views from the top and the visible region is 
mostly uniform across frames.

\begin{figure}[t]
\begin{center}
$\vimage{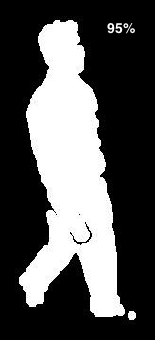}\hspace{2mm}
\vimage{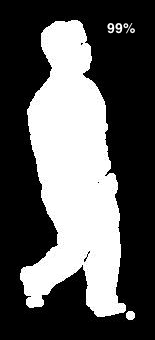}\hspace{2mm}
\vimage{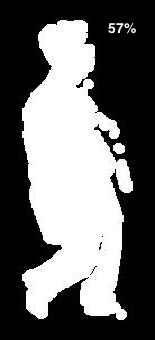}\hspace{2mm}
\vimage{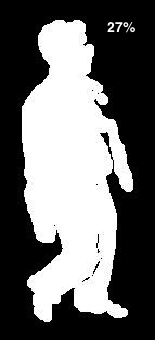}\hspace{2mm}
\vimage{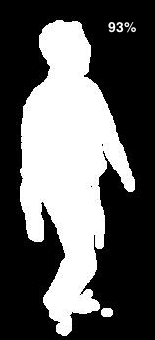}\hspace{2mm}
\vimage{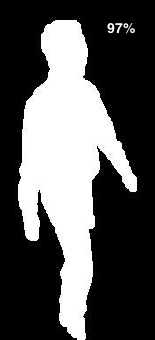}$
\caption{Example sequence with the predicted Bernoulli parameter printed.}
\label{fig_sequence}
\end{center}
\end{figure}

Next in Fig.~\ref{fig_sequence} we show a testing sequence with the predicted Bernoulli parameter $f_w(g_t;\theta_w)$ printed. 
After inspecting the Bernoulli parameter value on testing sequences, we observe large variations even among neighboring frames. 
Smaller values are typically associated with noisy frames, or frames with unusual pose (\emph{e.g.} person turning) and partial occlusions.

\begin{figure}[h]
\begin{minipage}{.47\linewidth}
  \centering
  \includegraphics[width=5.5cm]{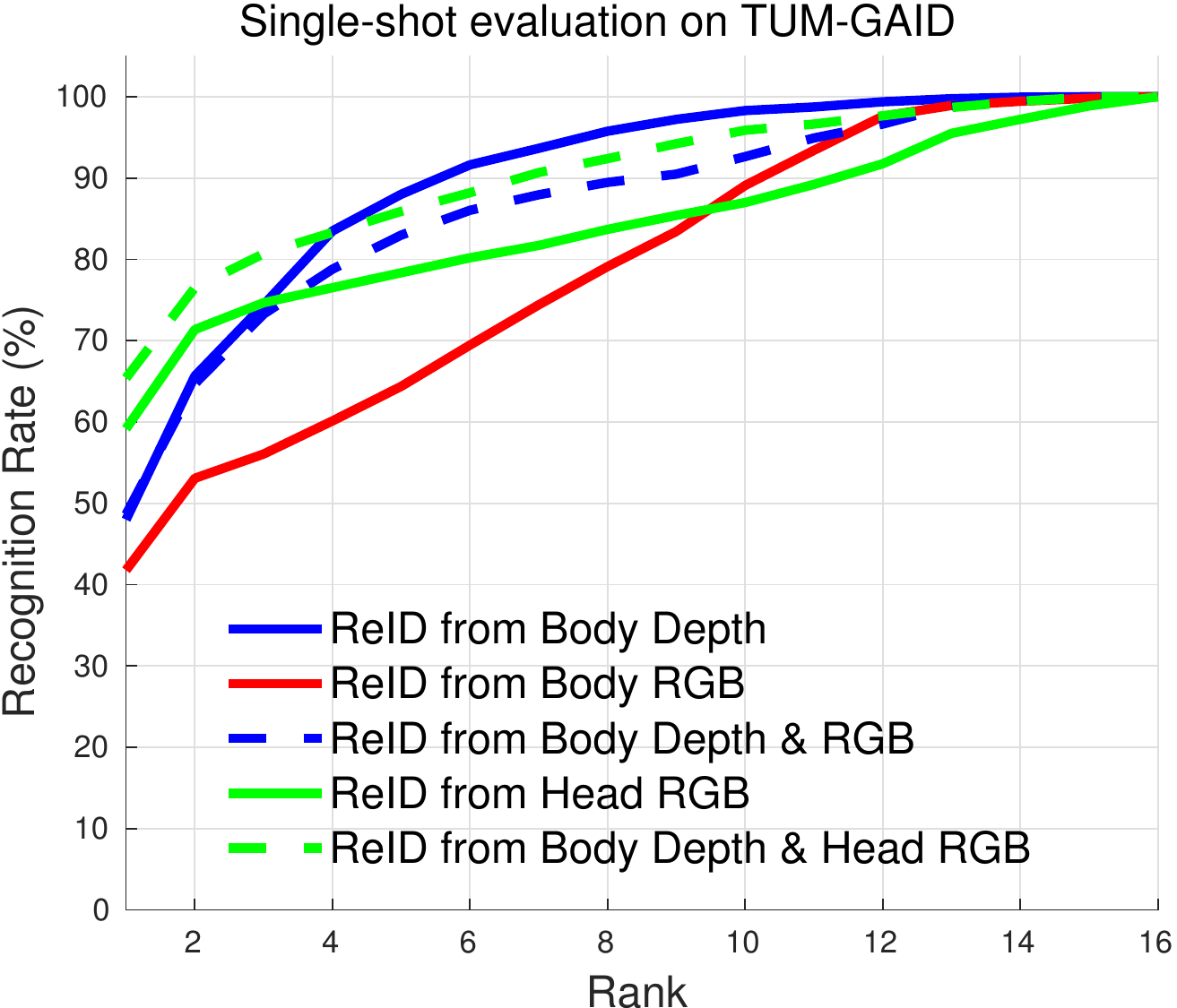}
  \vspace{0mm}
  \captionof{figure}{Cumulative Matching Curves (CMC) on TUM-GAID for the scenario that the individuals wear clothes which are not provided during training.}
\label{plot_tumgaid_task}
\end{minipage}%
\hfill
\begin{minipage}{0.5\linewidth}
\vspace{-2mm}
\centering
\captionof{table}{Top-1 re-identification accuracy (top-1, $\%$) and normalized Area Under the Curve (nAUC, $\%$) on TUM-GAID 
in new-clothes scenario with single-shot (\emph{ss}) and multi-shot (\emph{ms}) evaluation}
\vspace{2mm}
\resizebox{\linewidth}{!}{
\begin{tabular}{ c | c | c }
\hline Modality & top-$1$ & nAUC\\ \hline \hline
\emph{Body RGB (ss)~\cite{xiaoLOW16}} & 41.8 & 74.3 \\ \hline
\emph{Body Depth (ss)} & 48.0 & {\bf 85.0} \\ \hline
\emph{Body Depth} \& \emph{RGB (ss)} & 48.6 & 81.9 \\ \hline
\emph{Head RGB (ss)} & 59.4 & 79.5 \\ \hline
\emph{Body Depth} \& \emph{Head RGB (ss)} & {\bf 65.4} & 85.2 \\ \hline \hline
\emph{Body RGB (ms: LSTM} \& \emph{RTA)} & 50.0 & 79.9 \\ \hline
\emph{Body Depth (ms: LSTM)} & 56.3 & 87.7 \\ \hline
\emph{Body Depth (ms: LSTM} \& \emph{RTA)} & 59.4 & {\bf 89.6} \\ \hline
\emph{Head RGB (ms: LSTM} \& \emph{RTA)} & 65.6 & 81.0 \\ \hline
\begin{tabular}{@{}c@{}} \emph{Body Depth} \& \emph{Head RGB} \\ \emph{(ms: LSTM} \& \emph{RTA)}\end{tabular} & {\bf 75.0} & 88.1 \\ \hline
\end{tabular}}
\label{table_tumgaid}
\end{minipage}
\end{figure}

\subsection{Application in Scenario with Unseen Clothes}
\label{tumgaid}

Towards tackling our key motivation, we compare our system compared to a state-of-the-art RGB method in scenario 
where the individuals change clothes between the recordings for training and test set. We use the TUM-GAID database at which $305$ persons appear 
in sequences $n01$--$n06$ from session $1$, and $32$ among them appear with new clothes in sequences $n07$--$n12$ from session $2$.

Following the official protocol, we use the Training IDs to perform RGB-to-Depth transfer for our CNN embedding. We use sequences 
$n01$--$n04$, $n07$--$n10$ for training, and sequences $n05$--$n06$ and $n11$--$n12$ for validation. Next, we deploy the Testing IDs 
and use sequences $n01$--$n04$ for training, $n05$--$n06$ for validation and $n11$--$n12$ for testing. Thus, 
our framework has \emph{no access} to data from the session $2$ during training. However, we make the assumption that the $32$ subjects 
that participate in the second recording are known for all competing methods.

In Table~\ref{table_tumgaid} we show that re-identification from body depth is more robust than from body RGB~\cite{xiaoLOW16}, 
presenting $6.2\%$ higher top-1 accuracy and $10.7\%$ larger nAUC in single-shot mode. 
Next, we explore the benefit of using head information, which is less sensitive than clothes to day-by-day changes. 
To that end, we transfer the RGB-based pre-trained model from~\cite{xiaoLOW16} and fine-tune 
on the upper body part, which we call ``Head RGB". 
This results in increased accuracy, individually and jointly with body depth. 
Finally, we show the mutual benefits in multi-shot performance for both body depth, head RGB and their linear combination in class posterior. 
In Fig.~\ref{plot_tumgaid_task} we visualize the CMC curves for single-shot setting. We observe that ReID from body depth scales better 
than its counterparts, which is validated by the nAUC scores. 

\section{Conclusion}
\label{conclusion}

In the paper, we present a novel approach for depth-based person re-identification. To address the data scarcity problem, 
we propose split-rate RGB-depth transfer to effectively leverage pre-trained models from large RGB data 
and learn strong frame-level features. To enhance re-identification from video sequences, we propose the 
Reinforced Temporal Attention unit, which lies on top of the frame-level features and is not dependent on the network architecture. 
Extensive experiments show that our approach outperforms the state of the art in depth-based person re-identification, and 
it is more effective than its RGB-based counterpart in a scenario where the persons change clothes.

\paragraph{Acknowledgments:} This work was supported in part by ARO W911NF-15-1- 0564/66731-CS, ONR N00014-13-1-034, 
and AFOSR FA9550-15-1-0229.

%
%
%
\nocite{*}

\bibliographystyle{splncs04}

\end{document}